\documentclass{article}

\usepackage{arxiv}

\usepackage[utf8]{inputenc} 
\usepackage[T1]{fontenc}    
\usepackage{hyperref}       
\usepackage{url}            
\usepackage{booktabs}       
\usepackage{amsfonts}       
\usepackage{nicefrac}       
\usepackage{microtype}      
\usepackage{lipsum}		
\usepackage{graphicx}
\usepackage{natbib}
\usepackage{doi}
\usepackage{subfigure}
\usepackage{makecell}
\usepackage{appendix}

\title{AI and the Net-Zero Journey: Energy Demand, Emissions, and the Potential for Transition}

\author{{Pandu Devarakota}\textsuperscript{1} \\
        \And 
        {Nicolas Tsesmetzis}\textsuperscript{2} \\
        \And
        {Faruk O. Alpak}\textsuperscript{2} \\
        \And 
        {Apurva Gala}\textsuperscript{1} \\
        \And 
        {Detlef Hohl}\textsuperscript{1} \\ 
        \textsuperscript{1}{Shell Information Technology International Inc., Houston, TX - 77082, USA, Email: pandu.devarakota@shell.com}\\
        \textsuperscript{2}Shell International Exploration and Production Inc., Houston, TX - 77082, USA.
        }

\date{}

\hypersetup{
pdftitle={A template for the arxiv style},
pdfsubject={q-bio.NC, q-bio.QM},
pdfauthor={David S.~Hippocampus, Elias D.~Striatum},
pdfkeywords={First keyword, Second keyword, More},
}

\begin{document}
\maketitle

\begin{abstract}
	Thanks to the availability of massive amounts of data, computing resources, and advanced algorithms, AI has entered nearly every sector. This has sparked significant investment and interest, particularly in building data centers with the necessary hardware and software to develop and operate AI models and AI-based workflows. In this technical review article, we present energy consumption scenarios of data centers and impact on GHG emissions, considering both near-term projections (up to 2030) and long-term outlook (2035 and beyond). We address the quintessential question of whether AI will have a net positive, neutral, or negative impact on CO\textsubscript{2} emissions by 2035. Additionally, we discuss AI's potential to automate, create efficient and disruptive workflows across various fields related to energy production, supply and consumption. In the near-term scenario, the growing demand for AI will likely strain computing resources, lead to increase in electricity consumption and therefore associated CO\textsubscript{2} emissions. This is due to the power-hungry nature of big data centers and the requirements for training and running of large and complex AI models, as well as the penetration of AI assistant search and applications for public use. However, the long-term outlook could be more promising. AI has the potential to be a game-changer in CO\textsubscript{2} reduction. Its ability to further automate and optimize processes across industries, from energy production to logistics, could significantly decrease our carbon footprint. This positive impact is anticipated to outweigh the initial emissions bump, creating value for businesses and society in areas where traditional solutions have fallen short. In essence, AI might cause some initial growing pains for the environment, but it has the potential to support climate mitigation efforts.
\end{abstract}


\section*{Impact Statement}
The emergence of new technologies inevitably brings inherent challenges and a degree of skepticism regarding claimed benefits and practical applications. This is a natural part of technological evolution. In this article, we focus mainly on the impact of growing infrastructure needs for AI, specifically data centers, and the increasing energy demand required to power them. While the energy mix that will supply these data centers is important and is expected to evolve in the coming years, our main emphasis is on  CO\textsubscript{2} emissions and their broader climate impact. We will especially explore the interplay between the rise of AI, data centers, energy demand, and emissions. In determining whether AI results in a net positive or negative contribution to emissions, it's essential to consider scenarios both with and without dedicated emissions reduction measures. Additionally, we consider the carbon benefits directly attributed to AI across various sectors of energy industry. Quantifying these elements presents significant challenges and uncertainties, given the wide variation in estimates and forecasts from both public and private institutions. Nevertheless, we have made reasonable assumptions and considered existing paradigms, such as the potential for AI to reduce an organization's GHG emissions or achieve global CO\textsubscript{2} reductions presented in [\cite{bcg_climate_2022}]. Our analysis mainly draws on the energy scenarios presented in [\cite{Shell_Sce_2025}], which outlines global energy mix trends and the anticipated impact of technology on reducing emissions.  

\section*{Disclaimer}
This document presents scenario-based analysis and projections related to artificial intelligence (AI), energy demand, and CO\textsubscript{2} emissions. The findings and views expressed are exploratory and do not constitute forecasts, commitments, or guarantees by Shell or its affiliates. All forward-looking statements are subject to significant uncertainties and assumptions. Actual outcomes may differ materially. References to third-party data, technologies, or companies are for illustrative purposes only and do not imply endorsement or verification. This document should not be interpreted as a reflection of Shell’s current or future business strategy, investment decisions, or regulatory positions.

\section{Introduction}
The concept of net-zero greenhouse gas (GHG) emissions is crucial for combating climate change and limiting global warming to well below 2°C, ideally to 1.5°C pre-industrial levels, as outlined in the Paris Agreement [\cite{paris_agreement}]. Achieving net-zero means balancing the total amount of GHGs released into the atmosphere with an equivalent amount being removed, thus stopping the accumulation of these gases that drive climate change. This goal requires significant transformations across all sectors of the economy, including energy, transportation, manufacturing, and agriculture [\cite{Shell_Sce_2025}, \cite{DOE2024_AI4Energy}]. This necessitates a comprehensive strategy and a multifaceted approach involving significant change in energy systems, agriculture and land use, and in industrial processes. The sheer scale of this economic transformation is a major challenge [\cite{MckinseyReport2022}]. For example, to reach net zero by 2050, Shell scenarios suggest that \$3-4 trillion of commercially viable investment in low-carbon energy is required each year up from current \$1.7 trillion a year investment [\cite{Shell_energy_2024}]. While these are large numbers, the difference from a reasonable baseline energy system investment profile is much smaller.

Generating electricity is a growing contributor to climate change, primarily because coal, oil, and gas are and will still be widely used compared to cleaner energy sources [\cite{Shell_Sce_2025}]. While electricity demand has been steady over the years, it has recently surged and is projected to continue increasing its share until 2050. This growth is driven by the expanding digital economy and the rise of artificial intelligence (AI). Projections indicate a substantial surge in power consumption from data centers, which are essential for AI infrastructure. Some forecasts suggest that electricity demand from this sector could double by 2026 [\cite{iea2024}]. The increasing complexity and computational intensity of AI models, especially generative AI, require significantly more energy per query compared to traditional internet services. For instance, a single ChatGPT query consumes approximately ten times the electricity of a typical Google search [\cite{deVries2023}]. Although the numbers these estimates are based have already been outdated, and the AI assistants can generate more information than just word tokens, it is extensively quoted and generally accepted that queries to AI models consume more energy than simple Google searches. This rising demand puts considerable pressure on existing electricity infrastructure, necessitating a re-evaluation of energy production and consumption patterns [\cite{duke2025, epri_2024}].

Major tech companies and cloud providers like Microsoft and Google have reported that their emissions have increased recently [\cite{googlereport2024, microsoftreport2024}]. This trend is expected to continue due to the rush towards building more data centers in the next few years to support the rising AI demand and the ambition to become dominant players in the AI. Several reports, including the IEA report on AI and power demand trends, indicate that GHG emissions will inadvertently rise in the near future as the alternative energy sources are expected to take more time to become significant contributors [\cite{Shell_Sce_2025}].

However, AI also presents a transformative opportunity for accelerating the transition to a net-zero future. AI's unique ability to process vast amounts of complex data on energy systems, emissions, and climate impact allows for more informed and data-driven approaches to decarbonization. AI can optimize energy grids, improve the efficiency of renewable energy sources through better forecasting, and support the development of more sustainable industrial processes and materials. Additionally, AI can enhance energy efficiency, availability and reduce waste across various sectors, from transportation and buildings to industrial manufacturing, significantly contributing to emissions reduction. Some reports suggest that AI alone could potentially reduce global GHG emissions by 4\% to 16\% by 2030 [\cite{Shell_Sce_2025, bcg_climate_2022}], and 1.4 gigatonnes by 2035 in the widespread adoption case [\cite{iea2025}].

The advancements in \textit{AI for science} (see, e.g., [\cite{chandra2024fourier}]) demonstrate that AI can significantly accelerate complex challenges such as weather modeling and prediction, climate modeling, and CCS storage modeling technologies, potentially by up to 100,000 times. At Shell, we have conducted projects investigating the role of AI in CCS within both traditional aquifers and land use. Our findings indicate that AI can greatly expedite the process of identifying suitable sites and accelerating project execution by several orders of magnitude. We anticipate that in the next 5-10 years, these areas will present significant potential and opportunities to offset the carbon emissions [\cite{Pawar2025AcceleratedCS}]. 

This paper explores the complex interplay between AI and the pursuit of net-zero emissions. Section~\ref{methodology} lays down the methodology of calculating the emissions. The analysis of the exponential growth in demand for critical resources required to support AI development and deployment, and its consequent impact on CO\textsubscript{2} emissions in the near-term scenario is presented in Sections~\ref{methodology}, \ref{aiboom}, and \ref{datacenter}. This analysis highlights a critical juncture which requires careful consideration, investment, and development in the immediate future. Subsequently, we explore how ongoing advancements, including efficiency improvements in AI hardware and software, the adoption of alternative energy resources, and potential technological breakthroughs accelerated or enabled by AI, can reshape this trajectory in the long-term (Sections~\ref{consumption}, \ref{aievil} and \ref{ailongterm}).

\section{Methodology} \label{methodology}
To accurately estimate CO\textsubscript{2} emissions from data centers and AI workloads, a refined methodology is necessary. This involves moving beyond simple estimations of data center energy consumption to utilizing more granular data and applying location-specific, time-varying emission factors that explicitly incorporate uncertainty. Near-term scenarios should quantify the growth of data centers and AI workloads with defined probability distributions, reflecting the inherent unpredictability. Long-term projections need to model potential efficiency gains in AI hardware and software, the indirect impact of AI in optimizing energy systems and increasing renewable penetration, and the uncertain trajectory of the energy mix.

By employing Monte Carlo simulation with these probabilistically defined inputs, we can generate a range of potential emission scenarios for both the near and long term. This approach allows for a more nuanced understanding of the uncertainties involved and enables sensitivity analysis based on the factors with the most significant influence on future data center and AI-related emissions. The ultimate goal is to provide robust, probabilistic insights that can inform strategies for mitigating the environmental impact of this rapidly growing sector.

\section{The AI Boom and its Energy Appetite} \label{aiboom}

The AI landscape is experiencing unprecedented expansion, particularly in the areas of deep learning and large language models (LLMs). This surge in AI capabilities has been driven by breakthroughs in algorithms, the availability of large datasets, and advancements in computing hardware. Since the advent of ChatGPT, the capabilities of these LLMs have evolved from simple text generation and summarization to advanced reasoning abilities with remarkable accuracy and acceptance. These models are now routinely used on desktops (e.g., Copilot) and for content generation by creative artists. The bulk of these AI models are termed foundational models, which are trained on large datasets and are versatile and suitable for numerous downstream applications. GPT and Llama are both examples of these foundational models.

\begin{figure}[ht]
\vskip 0.2in
\begin{center}
    \centerline{\includegraphics[width=0.9\columnwidth]{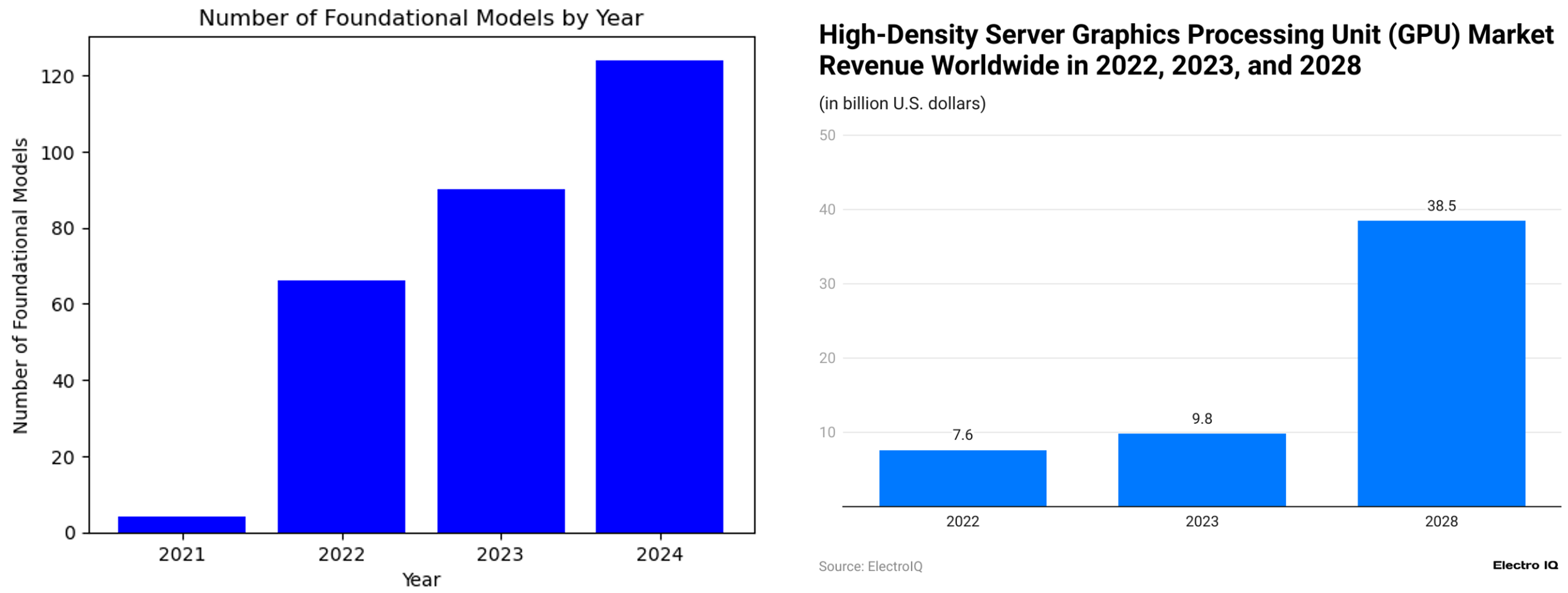}}
    \caption{The figure illustrates the growth of foundational models between 2021 and 2024, and the revenue growth of high-density GPU markets. Data courtesy from \cite{ecosystemgraph}}
\label{fig:foundational}
\end{center}
\vskip -0.2in
\end{figure}

Figure \ref{fig:foundational} shows the number of foundational models with more than 10 billion parameters built between 2021 and 2024 [data courtesy \cite{ecosystemgraph}].  It also illustrates the increase in revenues from high-density GPUs from 2022 and the projection for 2028 (Source: \cite{electroiq}). As can be seen, more investments are being made towards building GPU servers to support the development of large AI models. Although NVIDIA GPUs dominate the market for chips and low-level software libraries, there are no barriers for new entrants to join the arena of building new foundational models. Due to supply and demand mismatches and other approval process-related issues, commissioning grid infrastructure has temporarily slowed the pace of data center growth. However, the overall trend remains one of rapid expansion.

\begin{figure}[ht]
\vskip 0.2in
\begin{center}
    \centerline{\includegraphics[width=0.9\columnwidth]{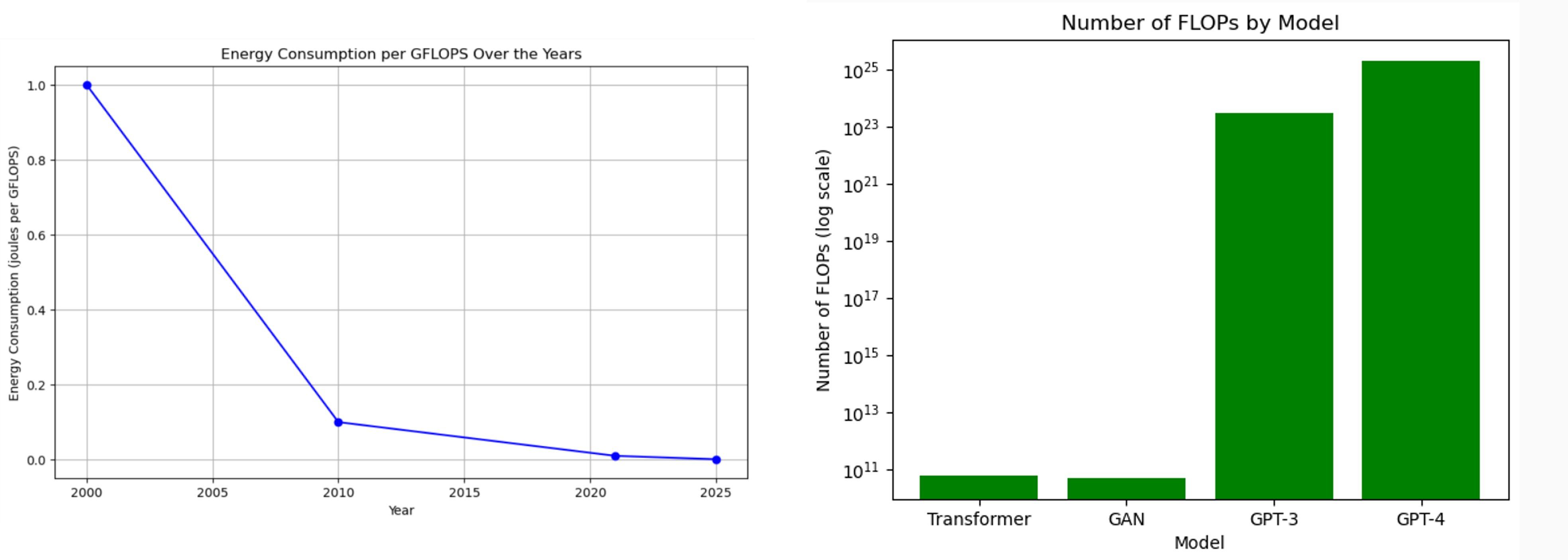}}
    \caption{The figure illustrates the decrease in energy consumption per GFLOPS over the years (left) and the number of FLOPS required to train different models (right)}
\label{fig:flops}
\end{center}
\vskip -0.2in
\end{figure}

However, this progress comes with significant consequences: a rapidly increasing demand for computational resources required to support the tasks in the life cycle of AI models, which involves both training (a one-time or occasional effort) and inference (repeatedly executed when the model is deployed for consumption). Although the work done (measured in FLOPs) per dollar and per watt has increased, meaning the power consumption per FLOP has decreased (see Figure~\ref{fig:flops} (left)), the number of FLOPS required to train AI models has continued to rise (see Figure~\ref{fig:flops} (right)). Consequently, the total power consumption for training AI models keeps increasing. This translates directly into escalating energy consumption [\cite{goldman2025}]. For instance, it was reported in [\cite{PattersonGoogle2021}] that training the large language model (LLM) powering ChatGPT-3 consumed almost 1300 megawatt hours of energy, equivalent to the annual energy usage of 130 American homes. Furthermore, an analysis by OpenAI suggests that the amount of power needed to train AI models has grown exponentially since 2012, doubling roughly every 3.4 months as the models become more sophisticated [\cite{openaiblog}].
\begin{figure}[ht]
\vskip 0.2in
\begin{center}
    \centerline{\includegraphics[width=0.9\columnwidth]{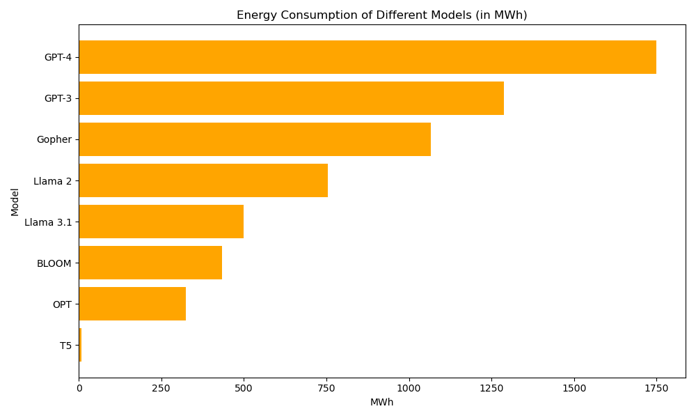}}
    \caption{The figure illustrates the electricity consumption of several prominent large language models published in recent years. Although the authors usually do not provide these statistics directly, some do. Many factors influence electricity consumption, but typically, if they disclose the hardware details—such as the number of GPU hours used and the types of GPUs—it's possible to calculate the overall power consumption using the GPUs' wattage. However, this calculation does not account for other components like CPUs, data storage, and cooling technology. Nonetheless, it gives an indication of the resources required to train these models.}
\label{fig:training}
\end{center}
\vskip -0.2in
\end{figure}

\textbf{Computational Demands of Training and Inference}: AI models, especially deep learning models and LLMs, require immense computational power for both training and inference [\cite{deVries2023}].
\begin{itemize}
    \item \textbf{Training} involves feeding the model vast amounts of data and adjusting its parameters to achieve the desired level of accuracy. This process can take weeks or even months and accounts for a significant portion (30-40\%) of an AI model's total electricity consumption (see Figure \ref{fig:training} for electricity consumption of training several prominent LLMs).
    \item \textbf{Inference} is the deployment and utilization of the trained AI models in real-world applications, where the model interprets new data and generates outcomes or predictions. Despite innovations in model efficiency, the inference phase accounts for the majority (60\%) of the energy footprint [\cite{epri_2024}] (see Table \ref{tab:inference} for electricity consumption of various tasks that AI chatbots are typically asked to execute).
\end{itemize}
While the trends clearly indicate a rise in the computational resources required for AI models, understanding the global consumption trends for all AI-related workloads necessitates knowing where these models are being trained and hosted during the post-deployment phase. These details will be discussed in the next section.

\section{Data Centers: The Epicenter of AI's Energy Demand} \label{datacenter}
Data centers are the backbone of the AI ecosystem, serving as the primary energy consumers that power the computational workloads required for training and deploying AI models. These facilities house vast arrays of servers, storage devices, and networking equipment, all of which demand significant amounts of electricity to operate. The rapid expansion of AI, particularly deep learning and large language models, has led to an unprecedented surge in data center growth, intensifying concerns about their environmental impact [\cite{epri_2024}].

The data center industry is experiencing rapid growth worldwide, driven by the increasing demand for cloud computing, digital services, cryptocurrencies, and AI applications. The number of data centers is increasing, and their sizes are also expanding. As of March 2024, there were approximately 9580 data centers globally, with more than half of them located in the United States [\cite{datacentermap}]. Between 2015 and 2024, the electricity consumption at data centers increased from under 200 TWh to over 400 TWh [\cite{iea2025}].  Data centers are projected to consume between 4.6\% and 9.1\% of U.S. electricity generation annually by 2030, up from an estimated 4\% today. Some forecasts even predicted that data centers could draw up to 21\% of the world's electricity supply by 2030 [\cite{Jones2018}]. In the US, fifteen states account for 80 percent of the national data center load, with data centers comprising a quarter of Virginia’s electric load in 2023. New data centers are being built with capacities ranging from 100 to 1000 megawatts, equivalent to the load from 80,000 to 800,000 homes. 

In [\cite{epri_2024}], the authors have developed four growth scenarios for US data center electricity load from 2023 to 2030: low growth (3.7 percent annual), moderate growth (5 percent annual), high growth (10 percent annual), and higher growth (15 percent annual). In the higher growth scenario, data center loads are projected to increase by a staggering 166\% from 152 TWh/year in 2023 to 403 TWh/year in 2030, with the majority of this demand attributed to AI. Such projections highlight the significant energy demands of AI and data centers and the need for innovative energy management and infrastructure planning to accommodate this growth. 

Compared to traditional data center infrastructure, the AI-focused data centers have significantly different needs and more energy demands. Hyperscale data centers, often needed for AI, have a minimum of 5,000 servers and at least 10,000 sq ft of space, providing extreme scalability. AI-ready data centers use accelerators like GPUs, NPUs, and TPUs to speed up machine learning and deep learning tasks. They consume massive amounts of electrical power and need advanced cooling systems to prevent outages and downtime. Liquid cooling offers greater efficiency in handling high-density heat, improving power usage effectiveness (PUE). These data centers also need high-speed networking and low-latency interconnects to facilitate rapid data movement between processors. Additionally, they use specialized storage and data management tools for handling unstructured data, such as NVMe SSDs and Lustre file systems. Designed for rapid scaling, AI data centers enable organizations to quickly deploy more computing power as AI workloads grow. However, they are more expensive due to the high cost of GPUs, specialized cooling, and high-speed networking components. 

Despite the growing emphasis on renewable energy, many regions still rely heavily on fossil fuels, especially, natural gas, to power their electricity grids [\cite{mittechreview2025}]. This may mean that a significant portion of data centers' energy consumption is directly contributing to GHG emissions. In regions with limited access to clean energy sources, the increasing electricity demands of data centers can exacerbate the reliance on fossil fuels, hindering efforts to decarbonize the energy sector. For instance, in [\cite{mittechreview2025}], the author quoted that natural gas is the only affordable choice given the need to meet the 24-7 electricity demand from the huge data center.

\subsection*{\textit{Specific Energy Needs of AI Workloads}}

As AI has become increasingly pervasive, it is difficult to estimate how much electricity consumption can be attributed solely to AI-related workloads. Some studies using proxies estimate that AI accounts for 10\% to 20\% of total data center electricity consumption [see \cite{iea2025} for example], but this percentage is rapidly increasing. AI workloads, including computation, storage, and networking, have specific energy needs within data centers [\cite{epri_2024, iea2024}].
\begin{itemize}
    \item \textbf{Computation:} AI model training and inference require immense computational power, driving up electricity consumption. The energy intensity of AI queries is significantly higher than traditional internet searches. For example, a ChatGPT request is estimated to require ten times the electricity of a typical Google query. See Section \ref{aiboom} for more detailed discussion on this. Studies indicate that computation accounts for 30-40\% of total energy consumption in data centers.  
    \item \textbf{Storage:} AI models often rely on vast datasets, necessitating large-scale storage infrastructure within data centers. Storing and accessing these datasets consumes additional energy.
    \item \textbf{Networking:} AI applications frequently involve data transfer between different servers and locations, placing demands on data center networking infrastructure. The energy consumption associated with data transmission contributes to the overall energy footprint.
    \item \textbf{Cooling systems:} These systems are critical for maintaining optimal temperatures within data centers to prevent hardware malfunction and ensure longevity. However, cooling is also one of the most energy-intensive aspects of data center operations, typically accounting for 30-40\% of total energy consumption. 
\end{itemize}

As we understand the global trends in data center growth and the rise of AI workloads at these facilities, along with the overall energy consumption patterns, we will present the near-term carbon footprint of AI and discuss its implications for achieving net-zero emissions in the next section.

\section{The Energy Consumption Scenarios and Near-Term Outlook for Data Centers and AI} \label{consumption}

\textit{Shell’s Energy Security Scenarios and the role of AI}

\begin{figure}[ht]
\vskip 0.2in
\begin{center}
    \centerline{\includegraphics[width=0.9\columnwidth]{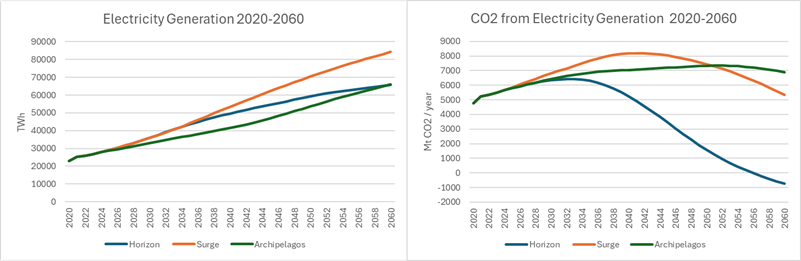}}
    \caption{Electricity generation and associated CO\textsubscript{2} emissions between 2020-2060 under the three Shell Energy Security scenarios. Of those, only \textit{Horizon} is reaching Net Zero by the 2050s. [Courtesy of \cite{Shell_Sce_2025}].}
\label{fig:delectricitygeneration_co2emissions}
\end{center}
\vskip -0.4in
\end{figure}

Shell’s 2025 Energy Security Scenarios explore AI’s transformative role across different pathways, each highlighting distinct impact on energy systems and societal change [\cite{Shell_Sce_2025}]. In the \textit{Surge} scenario, AI is a key driver of economic growth and societal transformation. AI optimizes manufacturing processes, enhances productivity across various sectors, and leads to increased economic activity. This surge in economic activity results in higher energy demand, which is managed efficiently by AI-powered smart grids. Additionally, AI accelerates the transition to low-carbon energy sources by optimizing the integration of renewable energy and enhancing energy storage solutions. The \textit{Archipelagos} scenario presents a different picture, where AI development is hindered by global concerns about resource security, border control, and trade restrictions. Geopolitical tensions limit the sharing of AI technologies and data, slowing progress. As a result, technological advancements are slower, impacting the transformation of energy systems. In the \textit{Horizon} scenario, AI plays a crucial role in achieving net-zero emissions by 2050. AI-driven innovations in energy efficiency and carbon capture technologies are pivotal in reaching this goal. AI also helps limit global warming to 1.5°C by optimizing energy consumption and reducing GHG emissions (Figure~\ref{fig:delectricitygeneration_co2emissions}).
Each scenario underscores the diverse roles AI can play in shaping future energy landscapes. The \textit{Surge} scenario emphasizes AI’s potential to drive rapid economic growth and energy demand, while the \textit{Archipelagos} scenario highlights the constraints and risks associated with geopolitical tensions. The \textit{Horizon} scenario focuses on AI’s role in achieving sustainability goals and mitigating climate change. 

\subsection*{\textit{Data Center Scenarios}}

\begin{figure}[ht]
\begin{center}
    \centerline{\includegraphics[width=0.99\columnwidth]{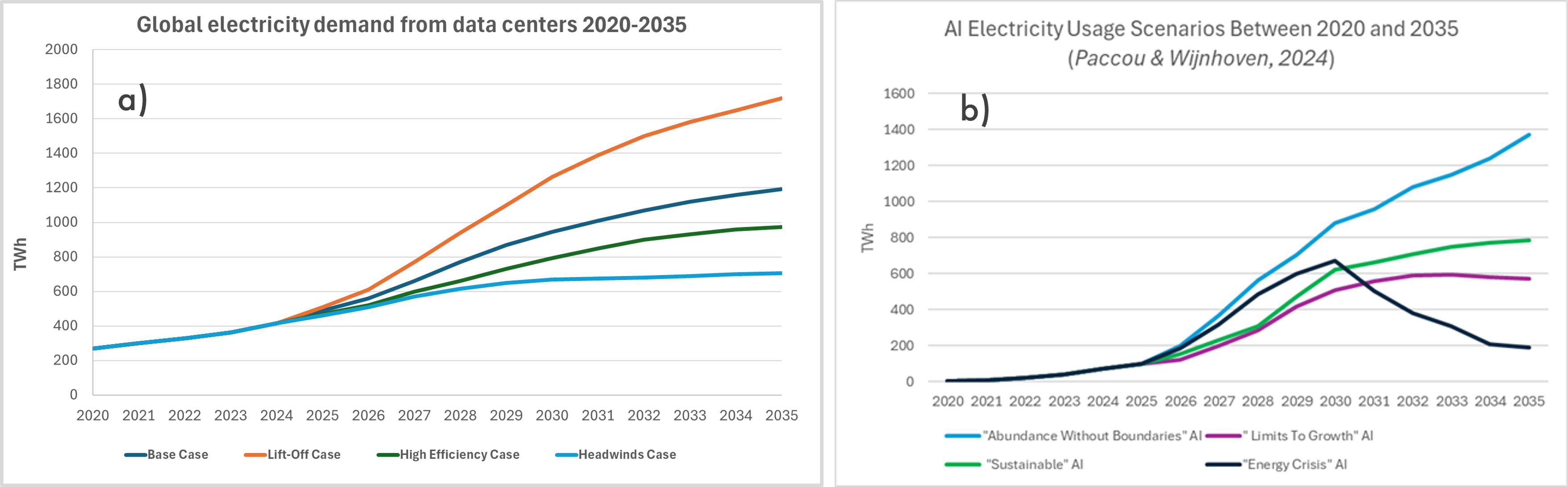}}
    \caption{a) Four scenarios for electricity demand by data centers globally between 2020-2035 (reproduced from \cite{iea2025}); b) Plausible scenarios for electricity usage by AI applications between 2020 and 2035. (reproduced from \cite{paccouandwijnhoven2024}).}
\label{fig:fourscenarios}
\end{center}
\vskip -0.4in
\end{figure}

Building on the insights from Shell’s Energy Securities scenarios, the Energy and AI 2025 report by the International Energy Agency (IEA) [\cite{iea2025}] outlines four sensitivity cases to explore different scenarios for AI uptake and its impact on the energy sector (Figure~\ref{fig:fourscenarios}a). The \textit{Lift-Off Case} assumes higher rates of AI adoption and proactive measures to reduce energy sector bottlenecks, envisioning rapid expansion of AI technologies and infrastructure. This scenario highlights the transformative potential of AI when supported by robust energy policies and investments, leading to significant advancements in AI capabilities and widespread adoption. The \textit{Headwinds Case} incorporates various bottlenecks, including macroeconomic challenges and delays in building out necessary energy infrastructure, representing a more cautious and constrained growth scenario. This case underscores the importance of addressing external factors that can hinder AI development, such as economic instability and infrastructure limitations, which can slow progress and reduce the overall impact of AI. 

The \textit{High Efficiency Case} focuses on significant improvements in energy efficiency, both in AI technologies and in the broader energy sector, assuming technological advancements will lead to reduced energy consumption per unit of AI output. This scenario emphasizes the potential for AI to drive sustainability and efficiency, reducing the environmental footprint of AI operations and enhancing the competitiveness of the energy sector. Lastly, the \textit{Baseline Case} serves as the reference scenario, assuming current trends and policies continue without significant changes, providing a benchmark to compare the other cases against. This case offers insights into the potential future of AI development if no major interventions are made, highlighting the risks and opportunities associated with maintaining the status quo. These four scenarios provide a broad view of potential futures, helping to understand the implications of AI development on energy consumption and sustainability.

\subsection*{\textit{AI Scenarios}}

In a recent study [\cite{paccouandwijnhoven2024}], the authors presented a comprehensive outlook using a system dynamics approach to model four scenarios for AI electricity consumption: \textit{Sustainable AI, Limits To Growth, Abundance Without Boundaries}, and \textit{Energy Crisis}. These scenarios provide insights into potential future trajectories of AI electricity use, influenced by current AI infrastructure decisions. While all scenarios show initial growth, their paths diverge after 2030 due to various internal and external factors (Figure~\ref{fig:fourscenarios}b). The \textit{Sustainable AI scenario} prioritizes efficiency and sustainability, with energy consumption increasing from 100 TWh in 2025 to 785 TWh in 2035. This scenario balances technological advancement with environmental stewardship. In contrast, the \textit{Limits To Growth} scenario shows minimal increase in consumption, from 510 TWh in 2030 to 570 TWh in 2035, indicating stunted economic expansion due to constraints like power availability and data scarcity. The \textit{Abundance With- out Boundaries} scenario projects unchecked growth, with consumption reaching 1,370 TWh by 2035. This scenario highlights risks such as power centralization and inequitable AI access. Meanwhile, the \textit{Energy Crisis} scenario underscores the risks of mismatched energy demand and infrastructure, with consumption peaking at 670 TWh before dropping to 190 TWh by 2035, indicating potential energy crisis.

Specific insights from these scenarios include the dominance of generative AI in the \textit{Sustainable AI} scenario, focusing on less energy-intensive models. The \textit{Limits To Growth} scenario highlights constraints in AI development due to power, chip manufacturing, and data scarcity. The \textit{Abundance Without Boundaries} scenario warns of risks associated with unchecked AI growth, including unsustainable infrastructure costs, global inequality, and increased e-waste. The \textit{Energy Crisis} scenario emphasizes the likelihood of underestimated future electricity needs due to insufficient grid planning and inaccurate demand projections. The authors note that projections for electricity consumption in TWh face challenges due to the nascent nature of the discipline, making direct comparisons difficult. However, some convergence is observed: the \textit{AI Abundance Without Boundaries} scenario projects 880 TWh usage by 2030, aligning with estimates by [\cite{pateletal2024}]. Another projection by [\cite{Kindig2024}] of 652 TWh by 2030 aligns with the \textit{Sustainable AI} scenario, suggesting a balanced growth trajectory. 

\begin{figure}[ht]
\begin{center}
    \centerline{\includegraphics[width=0.99\columnwidth]{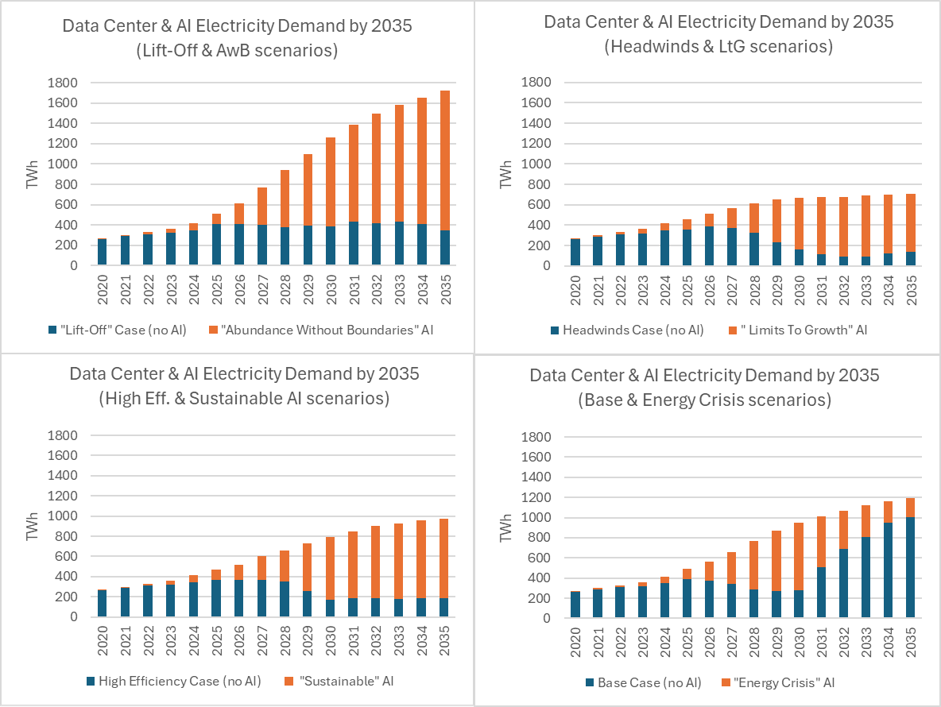}}
    \caption{Historical trends and forecasts for AI electricity demand through 2035. Depending on the scenario, it is estimated that electricity demand from data centers could range between 500 and 1,700 TWh, with AI workload expected to pick between 600 and 1,400 TWh at some point between 2030 and 2035.}
\label{fig:electricitydemand_trends_2035}
\end{center}
\vskip -0.4in
\end{figure}

To provide a comprehensive understanding of the potential futures for AI development and energy consumption, we have matched the Shell scenarios with the IEA sensitivity cases and the AI scenarios from \cite{paccouandwijnhoven2024}. This alignment is based on the shared themes and underlying assumptions of each scenario. The Shell scenarios offer a broad perspective on global economic and technological trends, while the IEA sensitivity cases focus on the specific impacts of AI on the energy sector. The AI scenarios from \cite{paccouandwijnhoven2024} provide detailed insights into the potential trajectories of AI electricity consumption. By matching these scenarios, we can explore how different factors such as geopolitical tensions, technological advancements, and sustainability efforts might influence the future of AI and energy. This integrated approach allows us to identify common challenges and opportunities, offering a more nuanced and holistic view of the interplay between AI development and energy consumption:

\begin{enumerate}
    \item \textbf{Surge - Lift-Off Case - Abundance Without Boundaries}: The \textit{Surge} scenario envisions rapid AI adoption and technological advancements, driving significant economic growth. Similarly, the \textit{Lift-Off Case} from the IEA assumes high AI adoption rates and proactive measures to reduce energy bottlenecks, highlighting AI's transformative potential. The \textit{Abundance Without Boundaries} scenario by \cite{paccouandwijnhoven2024} projects unrestricted AI growth, emphasizing its power to drive economic and technological progress without major constraints. These scenarios share a vision of rapid and expansive AI growth fueling economic and technological advancements. Under these scenarios, the energy demand by data centers continues to grow past 2035 exceeding 1700 TWh. AI accounts for more than three quarters of the electricity usage (Figure~\ref{fig:electricitydemand_trends_2035}a).  
    \item \textbf{Archipelagos - Headwinds Case - Limits To Growth}: The \textit{Archipelagos} scenario depicts technology development hindered by global concerns about resource, border, and trade security, leading to fragmented efforts and slower progress. Similarly, the \textit{Headwinds Case} from the IEA includes various bottlenecks, such as macroeconomic challenges and infrastructure delays, representing cautious growth with significant obstacles. The \textit{Limits To Growth} scenario by \cite{paccouandwijnhoven2024} highlights constraints on AI development due to resource limitations, emphasizing the need for careful management of AI's energy demands. These scenarios share a focus on limitations and challenges that hinder AI and energy sector progress. Under these scenarios, energy demand by data centers plateaus by 2030 at around 700 TWh. AI accounts for approximately 80\% of the electricity usage (Figure~\ref{fig:electricitydemand_trends_2035}b).  
    \item \textbf{Horizon - High Efficiency Case - Sustainable AI}: The \textit{Horizon} scenario focuses on achieving net-zero emissions by 2050 and limiting global warming to 1.5°C, emphasizing sustainability and efficiency in energy use. Similarly, the \textit{High Efficiency Case} from the IEA assumes significant improvements in energy efficiency, both in AI technologies and the broader energy sector, envisioning reduced energy consumption per unit of AI output. The \textit{Sustainable AI} scenario by \cite{paccouandwijnhoven2024} balances AI development with sustainability, highlighting the importance of generative AI inferencing and traditional AI in decarbonization efforts. These scenarios share an emphasis on achieving sustainability and efficiency in AI and energy consumption. Under these scenarios, energy demand by data centers approaching a plateau by 2035 at nearly 1000 TWh. AI accounts for approximately 80\% of the electricity usage (Figure~\ref{fig:electricitydemand_trends_2035}c).  
    \item \textbf{Baseline Case - Energy Crisis}: The \textit{Baseline Case} from the IEA assumes current trends and policies continue without significant changes, serving as a benchmark for comparing other scenarios and representing a future where AI development follows existing trajectories. The \textit{Energy Crisis} scenario by \cite{paccouandwijnhoven2024} considers the possibility of an energy crisis caused by AI, highlighting the risks of unmanaged AI development leading to significant energy consumption and potential shortages. It emphasizes the need for strategic planning and management to avoid such crises. These scenarios share a focus on the potential risks and challenges associated with AI development and energy consumption. Under these scenarios, energy demand by data centers approaching a plateau in 2035 at around 1200 TWh. AI electricity usage reaches its maximum in 2030 where it accounts for 70\% of the data center’s electricity usage. However, due to the energy crisis, it drops to around 16\% by 2035 (Figure~\ref{fig:electricitydemand_trends_2035}d).
\end{enumerate}

\section{The Near-Term Carbon Footprint: AI as a Net-Contributor} \label{aievil}
Although energy consumption scenarios for data centers are based on global estimates, they can greatly vary by country. This leads to a diverse global landscape influenced by global trends, local conditions, policies, and decisions. In fact, geopolitical factors play a crucial role in shaping AI’s impact on the energy transition and carbon footprint in the short term. Regulatory frameworks across different countries can either accelerate or hinder AI advancements, influencing how quickly AI technologies are adopted in energy sectors. National security concerns drive investments in AI for defense and surveillance, which can divert resources from sustainable energy initiatives. International trade policies and technological alliances also affect the global distribution of AI technologies, impacting their deployment in renewable energy projects [\cite{weforumpodcast2024}]. Control over semiconductor supply chains, essential for AI hardware, can create geopolitical tensions, affecting the availability and cost of AI technologies. Additionally, countries leading in AI development, such as the US and China, have significant influence over global AI policies and standards, shaping the direction of AI applications in energy efficiency and carbon reduction. These geopolitical dynamics, combined with societal factors like public perception and workforce adaptation, will determine AI’s role in achieving a sustainable energy transition and reducing carbon emissions [\cite{Iqbal2025}].

Despite the growing emphasis on renewable energy, many regions still rely heavily on fossil fuels to power their electricity grids. This means that a significant portion of data centers’ energy consumption is indirectly contributing to GHG emissions. In regions with limited access to clean energy sources, the increasing electricity demands of data centers can exacerbate the reliance on fossil fuels, hindering efforts to decarbonize the energy sector [\cite{bcg_climate_2022}].

\begin{figure}[ht]
\begin{center}
    \centerline{\includegraphics[width=0.9\columnwidth]{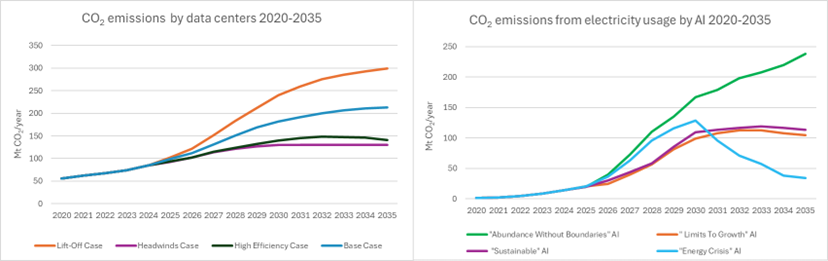}}
    \caption{Historical trends and forecasts for CO\textsubscript{2} emissions from a) data centers and b) AI workloads through 2035. These forecasts are based on the data provided in the \cite{Shell_Sce_2025}, \cite{iea2025} and \cite{paccouandwijnhoven2024}}
\label{fif:co2foecast_scenarios}
\end{center}
\vskip -0.4in
\end{figure}

As we probe the various scenarios in data center growth and the rise of AI workloads at these facilities, it will be important to deliberate their near-term carbon footprint and its implications for achieving net-zero emissions. Depending on the energy mix used to generate electricity, the CO\textsubscript{2} emissions largely vary from region to region. In some locations, the data centers claim to be 100\% renewable, but in many different places, it varies. Due to the lack of detailed CO\textsubscript{2} data for electricity generation in the three Shell scenarios, we based our calculations on the CO\textsubscript{2} emissions from the primary energy production (not only electricity). As the primary energy mix changes over time from fossil fuel to renewable energy, the associated CO\textsubscript{2} emissions are declining accordingly. As we see in the Figure~\ref{fif:co2foecast_scenarios}a, the rate of decline varies between scenarios. For the CO\textsubscript{2} estimates by data centers and AI workloads, the corresponding Shell scenarios were used as described in the previous section. For the \textit{Base Case} scenario, the average energy mix between the \textit{Surge} and \textit{Archipelagos} scenarios was used as they represent rather opposing cases. 

More precisely, the CO\textsubscript{2} emissions by data centers in the \textit{Headwinds Case} and the \textit{High Efficiency Case} reached plateau by 2030 at around 130-150 Mt CO\textsubscript{2} per year. In the \textit{High Efficiency Case} the trend started pointing downwards by 2035. Finally, in the \textit{Base} and \textit{Lift-Off Cases}, the CO\textsubscript{2} emissions continued to increase past 2035 albeit at different rates reaching 210 and 300 Mt per year respectively (Figure~\ref{fif:co2foecast_scenarios}a).  In the case of AI, the CO\textsubscript{2} emissions from electricity usage in \textit{Sustainable AI} and \textit{Limits to Growth} plateaued around 2030 at around 115 Mt per year. On the contrary, the \textit{Abundance without Boundaries} scenario grew at a much faster pace reaching 240 Mt CO\textsubscript{2} by 2035 with no clear signs of slowing down. Finally, the \textit{Energy Crisis} scenario peaked at 130 Mt CO\textsubscript{2} in 2030 followed by a rapid decline to 35 Mt within five years (Figure~\ref{fif:co2foecast_scenarios}b). For context, 100 Mt CO\textsubscript{2} is equivalent to the emissions from nearly 13.5 million American home’s energy usage for one year (\cite{epacalculator}).

As illustrated in Figure~\ref{fif:co2foecast_scenarios}, these scenarios present contrasting perspectives on the penetration of AI and its impact on CO\textsubscript{2} emissions. For example, the \textit{Lift-Off Case} from IEA is more optimistic, projecting the highest CO\textsubscript{2} emissions, whereas the \textit{Energy Crisis} scenario from \cite{paccouandwijnhoven2024} is pessimistic, predicting the least emissions. Although these reports present extreme scenarios, we believe the reality lies somewhere in between. It is often preferable to provide an interval within which the forecasted CO\textsubscript{2} emissions are expected to fall. To determine this interval, we consider the mean scenario presented in Figure~\ref{fif:co2foecast_scenarios} and calculate the P5, P50, and P95 probability cases around that, which will be described next.

\subsection*{Mean Scenario and Probabilistic Assessment}

To generate the potential range of CO\textsubscript{2} emissions under various assumptions and scenarios and to understand the variability and uncertainty in future emissions projections, a comprehensive methodology has been developed. This methodology includes a list of factors that influence emissions in data centers and the relationship between these factors and emissions. Table~\ref{tab:key_variables} describes the factors that we considered to influence the emissions at the data centers. It is important to note that the mean scenario calculated from Figure~\ref{fif:co2foecast_scenarios} is captured in the last column, and the values for other factors were taken from various reports and studies carried out that represent the actual numbers available until 2024 and the forecasts until 2035.

The energy mix factor and AI workload at the data centers used for Figure~\ref{fif:co2foecast_scenarios} have also been reported here. The mean and standard deviation for each variable are used to define the 99\% confidence interval, which provides the probabilistic range that the Monte Carlo (MC) simulations can use to generate many such possibilities and then calculate the emissions for each possibility. Our main hypothesis is that the extreme scenarios depicted in Figure~\ref{fif:co2foecast_scenarios} would be highly unlikely to occur. An important point to note here that these years include two pandemic years where there is no significant deviation in the variation for these factors.

In order to make predictions for each possibility, a mathematical model was fitted to the data presented in the Table~\ref{tab:key_variables} with columns 2 to 5 as inputs and column 6 as output. Historical data is integrated into the model to validate its accuracy and ensure that the projections are grounded in real-world observations. For example, the mean scenario curve honors perfectly the historical and forecasts data presented in Table~\ref{tab:key_variables}. Different scenarios (P5, P50, P95) represent varying levels of probability, with P5 being the pessimistic one with 5\% of the time the outcome will fall below this value, P50 the median, and P95 the optimistic one with 95\% of the time that the outcome will fall below this value. 

Key assumptions made in the methodology include the stability of certain factors over time and the exclusion of extreme events. These assumptions, along with potential limitations and uncertainties, are acknowledged to provide a balanced view of the projections. The impact of AI workloads on CO\textsubscript{2} emissions is specifically accounted for in the model, recognizing that AI-related activities can significantly influence energy consumption. The energy mix factor, which defines the ratio of fossil versus renewable energy used, is a critical component of the model, as it directly affects the emissions calculations.

Below is a summary of the methodology and steps involved to compute the P5, P50, P95 probability scenarios:
\begin{enumerate}
    \item Calculate the mean CO\textsubscript{2} emission scenario from all the scenarios presented in Figure~\ref{fif:co2foecast_scenarios}.
    \item Identify factors potentially impacting CO\textsubscript{2} emissions at data centers and gather past and future forecasts from published reports and literature from 2020 to 2035.
    \item Develop and validate a model honoring the data from the step 2.
    \item Determine the mean and standard deviation of each variable and the corresponding 99\% confidence interval for data variation.
    \item Perform MC simulations using combinations of parameters sampled from the confidence intervals and apply the approximated curve to find the corresponding emissions for each realization.
    \item Calculate the P5, P50, and P95 probability cases based on the predictions.
\end{enumerate}

\begin{figure}[ht]
    \centering
    \subfigure[]{
        \includegraphics[width=0.48\textwidth]{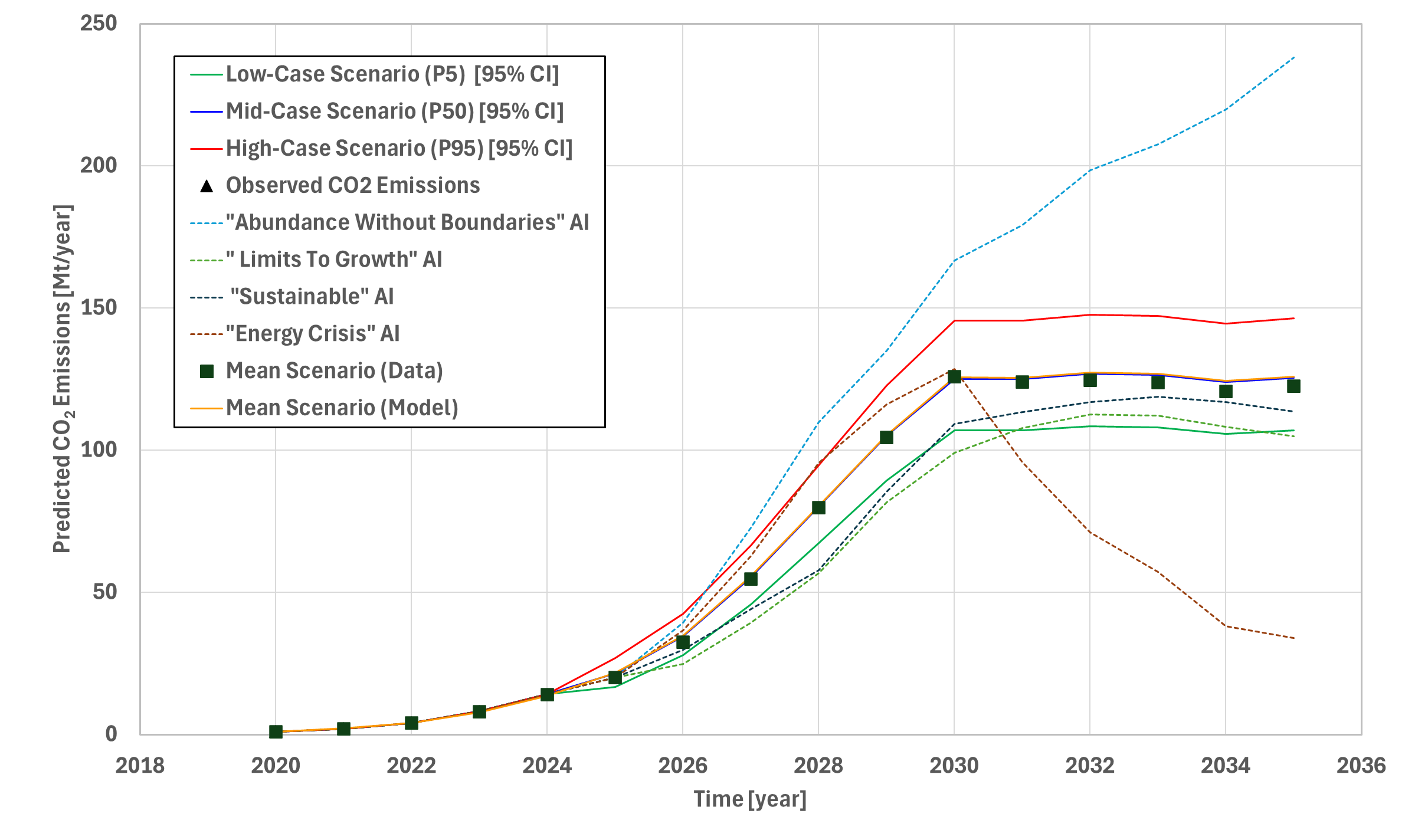}
        \label{fig:99ci}
    }
    \hfill
    \subfigure[]{
        \includegraphics[width=0.48\textwidth]{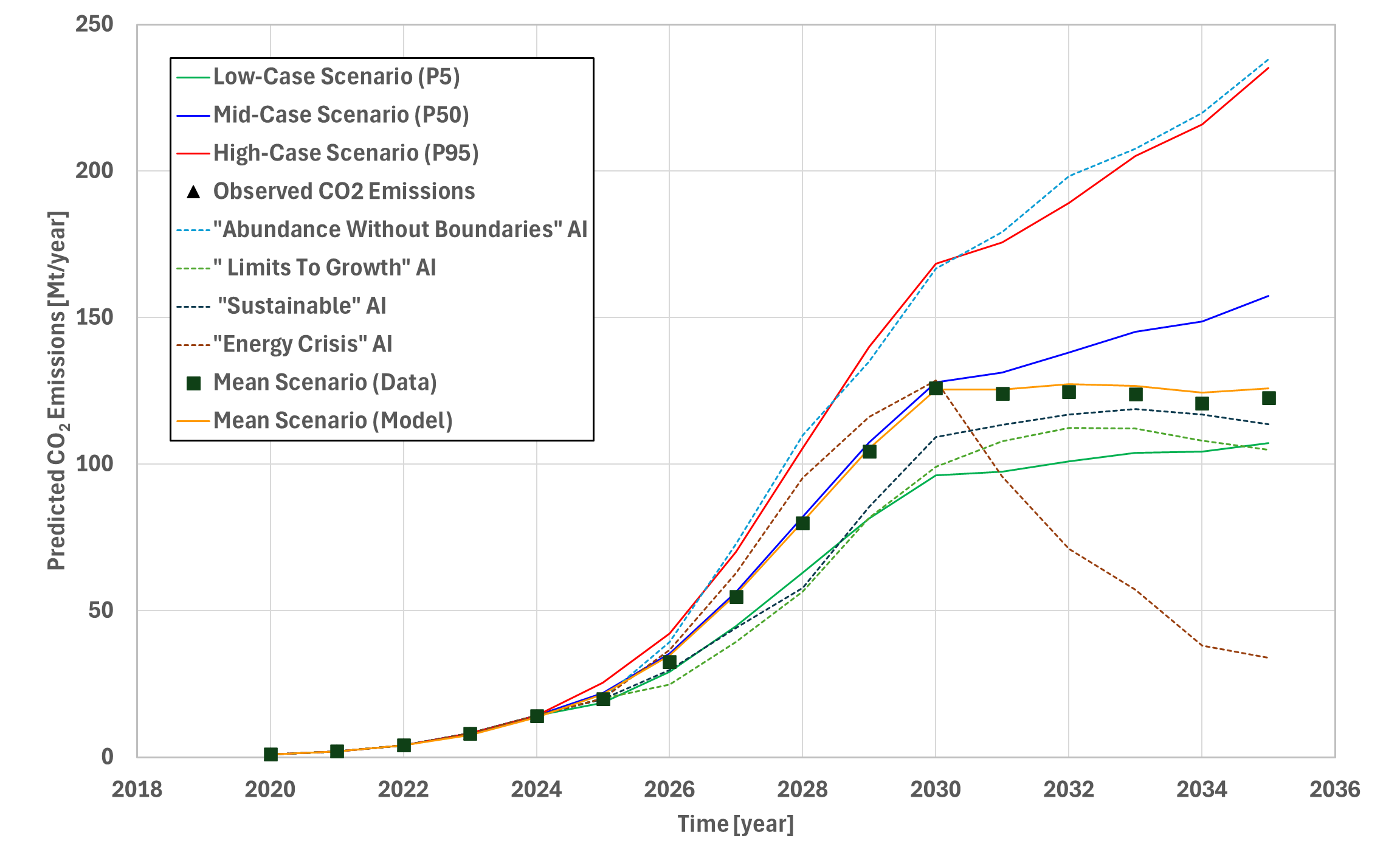}
        \label{fig:second_subfigure}
    }
    \caption{The P5, P50, P95 probability scenarios generated based on the following assumptions: a) The input variables listed in Table~\ref{tab:key_variables} were assumed to follow a normal distribution. The ranges for these variables were selected within a 99\% confidence interval around the mean scenario. b) The scenarios presented in \cite{paccouandwijnhoven2024} were reproduced without considering the energy crisis scenario.}
    \label{fig:99ci_noenergycrisis}
\end{figure}

Figure~\ref{fig:99ci_noenergycrisis} illustrates the P5, P50, and P95 probability scenarios calculated using the methodology described above. As it can be clearly depicted, the P5, P50, and P95 cases showcase a healthy deviation from the mean case; however, the deviation is not as extreme as predicted in either the IEA or Paccou et al. Our modeling does not prioritize an energy crisis scenario as the most probable outcome, though such risks cannot be ruled out. We expect a deviation from the mean scenario but not extreme ones. The reason is that most of the emissions are expected to come from data centers, and the data center growth does not support such an extreme scenario. On the other hand, we do not expect an energy crisis scenario, as we have not seen such a one in the last 20 years (including the COVID time).

In summary, while AI applications have clear benefits for environmental initiatives, it is challenging to translate these insights into actual carbon reduction efforts due to the time it takes for AI to have an impact on carbon reduction. Despite technical advances such as model optimizations (e.g., low-rank approximations, smaller and more efficient models), green AI hardware, and the use of renewable energy sources to power data centers, we anticipate that AI will have a net positive impact on carbon emissions in the near future. It will take several years, possibly beyond 2035, for AI to start having a net negative impact on carbon emissions due to technical advancements and penetration, which will be discussed in the next section.

\section{The Long-Term Promise: AI as a Net-Zero Enabler} \label{ailongterm}

The increasing presence of AI raises concerns about its energy consumption. However, it also has the potential to reduce emissions and help achieve climate targets. AI applications can optimize energy usage in sectors such as transportation, manufacturing, and agriculture. For example, AI-powered smart grids can enhance grid efficiency, minimize energy loss during transmission, and seamlessly integrate renewable energy sources. Studies indicate that intelligent-applications coupled with smart grids could reduce energy consumption by 15\% [\cite{Jin2024}]. AI-based applications are already enabling predictive maintenance, anticipating equipment failures, thereby reducing downtime and energy waste. Furthermore, AI can optimize supply chains, leading to decreased transportation emissions. AI can accelerate project delivery timelines for CCS through very fast surrogate-based modeling. It is difficult to gauge the overall impact of AI on CO\textsubscript{2} reductions as there are many technological, socio-economic, and political factors influencing that journey. A recent study from Google and BCG suggests that implementing AI for corporate sustainability could result in a substantial 2.6 to 5.3 gigaton reduction in CO$_2$ equivalent by 2030 if potential optimizations were to fully penetrate industry [\cite{bcg_climate_2022}]. Similarly, [\cite{iea2025}] reported that AI applications in the energy sector could lead to a reduction of 1.4 gigaton of CO\textsubscript{2} by 2035 in the Widespread Adoption Case. 

Construction of the energy system is shifting from large bespoke projects built in the field to modular units produced on assembly lines, for example solar photovoltaic modules and grid batteries. Another example is small modular reactors (SMRs), which are a new generation of nuclear reactors designed to be smaller, more flexible, and more cost-effective than traditional large-scale reactors. SMRs can be manufactured in factories and transported to sites, which reduces construction time and costs. They offer enhanced safety features, such as passive cooling systems that do not require human intervention or external power sources. Additionally, SMRs can be deployed in remote locations or integrated into existing power grids, making them a versatile option for expanding nuclear energy capacity and reducing carbon emissions. Developments in a number of decarbonization and carbon removal technologies are crucial to lower energy system emissions. AI has started accelerate and increase the efficiency of such technologies.

\textbf{Nuclear Power}: AI is aiding nuclear energy in reducing emissions through various applications. AI is being used in optimizing the designs and operation of small modular reactors (SMRs) and their integration into the energy grid. AI algorithms are finding ways to enhance the efficiency of SMRs by predicting the most effective operational parameters, thus maximizing energy output while minimizing waste. By optimizing reactor operations, AI enhances energy generation efficiency [\cite{app15020999}]. Predictive maintenance powered by AI helps foresee equipment failures, reducing downtime and ensuring peak operational efficiency [\cite{MYERS2025105568}]. Accurate energy demand forecasting allows for better planning, reducing reliance on backup fossil fuel plants [\cite{app15020999}]. AI also enhances safety by monitoring reactor data in real-time to identify anomalies, thus preventing potential accidents [\cite{MYERS2025105568}]. Additionally, AI improves grid management, ensuring efficient energy distribution and lowering fossil fuel dependence [\cite{yale2025}].

\textbf{Wind and Solar}: AI is playing a transformative role in reducing emissions from wind and solar energy by optimizing various aspects of these renewable energy systems. In wind energy, AI enhances efficiency through precise weather forecasting and wind analysis, allowing for better planning and operational adjustments. AI-driven predictive maintenance helps identify potential issues before they become significant problems, reducing downtime and ensuring turbines operate at peak efficiency. Additionally, AI-powered tools for turbine monitoring and inspection improve safety and performance by detecting defects that might be missed by human inspectors [\cite{dorterler2024}, \cite{morkos2025}]. In solar energy, AI maximizes energy output by analyzing data from various sources, such as weather conditions and system performance, to predict energy production and optimize energy storage. Predictive analytics and machine learning algorithms enhance the efficiency of solar panels by managing factors like shading and angle adjustments [\cite{GUL2025125210}, \cite{solarai}]. AI also plays a crucial role in improving grid management, ensuring that the energy generated from wind and solar sources is distributed efficiently, reducing reliance on fossil fuels and lowering overall emissions [\cite{yale2025}].

\textbf{Biofuels}: By leveraging AI and machine learning algorithms, researchers can optimize the entire biofuel production process, from feedstock selection to final fuel production. AI helps in analyzing large datasets to identify the most efficient feedstocks and processing conditions, thereby maximizing yield and minimizing environmental impact. Predictive models developed using AI can forecast the performance of different biofuel production methods, allowing for better planning and resource allocation [\cite{OKOLIE2024100928}]. Additionally, AI-driven automation in bioprocessing can enhance the efficiency and consistency of biofuel production, reducing waste and emissions. AI also aids in the development of advanced catalysts and enzymes that can accelerate biofuel production while lowering energy consumption [\cite{doe2023}, \cite{Salama2025}]. 

\textbf{Hydrogen}: Green hydrogen energy, produced through water electrolysis, has emerged as a crucial technology in this transition. Proton exchange membrane water electrolyzers (PEMWEs) have gained significant attention due to their efficiency and compatibility with renewable energy sources, positioning them as a frontrunner in green hydrogen production technologies. Challenges with PEMWEs include the need for innovative research to overcome existing limitations in key materials such as electrocatalysts, membranes, membrane electrode assemblies (MEAs), stacks, and systems. AI's transformative potential in  PEMWE research is focusing on discovery and optimization of key materials, as well as the design and management of MEAs, cells, and stacks [\cite{WANG2025}]. AI-enabled simulations provide valuable insights during feasibility studies, helping to determine the optimal number of electrolyzers, capacity utilization, and energy demand for hydrogen production. These simulations also incorporate renewable energy models, weather forecasts, and electricity market prices to optimize the entire lifecycle of hydrogen projects. Furthermore, AI plays a vital role in managing energy grids, ensuring efficient integration of hydrogen energy and reducing reliance on fossil fuels [\cite{MULLANU2024}]. 

\textbf{Carbon Capture and Storage (CCS)}: Several key technological advancements in CCS technologies that are crucial for reducing the carbon footprint of energy-intensive industries, including AI. One significant advancement is the development of smaller, more efficient capture systems. These systems are designed to be more compact and cost-effective, making them suitable for a wider range of applications. For example, the use of modular capture units that can be easily integrated into existing industrial processes. These units are capable of capturing CO\textsubscript{2} emissions at a lower cost and with higher efficiency compared to traditional systems. Another important development is the integration of CCS with renewable energy sources. This approach involves using renewable energy to power the capture and storage processes, thereby reducing the overall carbon footprint of the CCS operations. For instance, some CCS projects are now being powered by solar or wind energy, which not only reduces emissions but also enhances the sustainability of the entire process. New storage techniques, such as mineral carbonation and enhanced oil recovery (EOR), are being developed with the help of AI to improve the efficiency and safety of CO\textsubscript{2} storage. Recent advances in operator learning are accelerating flow simulations conducted to quantify the CO\textsubscript{2} plume migration by three to four orders of magnitude [\cite{Pawar2025AcceleratedCS}, \cite{chandra2024fourier}]. These technological advancements are crucial for reducing the carbon footprint of energy-intensive industries, including AI. By making CCS more efficient, cost-effective, and sustainable, these innovations help mitigate the environmental impact of AI and other advanced technologies, contributing to global climate goals [\cite{gsccs2024}].

\textbf{Quantum Computing}: Advances in quantum computing have the potential to significantly impact AI applications, reduce energy demand for data centers, and contribute to the reduction of overall GHG emissions. Quantum computing can enhance AI by providing unprecedented computational power, enabling more advanced models and algorithms, and optimizing AI processes for greater efficiency. This can lead to breakthroughs in natural language processing, optimization, and machine learning. In data centers, quantum computing can reduce energy consumption by performing calculations more efficiently and optimizing cooling processes through advanced quantum algorithms. This efficiency can lower operational costs and reduce the environmental impact of data centers. Additionally, quantum computing can aid in the development of innovative climate solutions, such as new materials and catalysts for carbon capture and storage, and more efficient renewable energy technologies. These advancements can play a crucial role in mitigating climate change and reducing GHG emissions, aligning with global sustainability goals. However, quantum computing is still in its nascent phases, and the realization of its promises is not expected anytime soon.

As part of the energy system decarbonization, AI can also transform transportation, a major source of emissions. Self-driving vehicles, optimized traffic systems, and improved logistics planning have the potential to significantly reduce our reliance on fossil fuels. By analyzing real-time data, AI can forecast traffic trends, streamline routes, and reduce congestion. Research on traffic prediction encompasses various models with examples including feature-based models [\cite{Li2019}], Gaussian process models [\cite{Salinas2019}], state-space models [\cite{Duan2019}], and deep learning models [\cite{Yin2022}]. Autonomous vehicle development is another nascent area for AI [\cite{bojarski2016},\cite{pmlr-v78-wu17a}] predominantly targeting safety and energy efficiency. AI-driven smart mobility systems has the potential to improve the energy efficiency of public transportation through dynamic scheduling to guarantee timely arrivals and departures. Finally. AI can accelerate the discovery of sustainable materials for vehicles and associated manufacturing processes. Another important element of the energy system is heating and cooling of buildings. AI models can process sensor data at the building and city level to optimize heating and cooling systems, lighting, and energy consumption patterns [\cite{Qolomany2019}, \cite{PHAM2020121082}, \cite{ZEKICSUSAC2021102074}, \cite{Ghazal2023}] leading to energy savings and reduction in carbon emissions [\cite{MILOJEVICDUPONT2021102526}, \cite{ZHANG2023101347}].

\section{Mitigating AI's Energy Impact: Strategies and Solutions} \label{mitigation}

Research in green AI hardware can transform the sustainability of artificial intelligence by prioritizing energy efficiency and reducing environmental impact. Recent advancements include energy-efficient Graphics Processing Units (GPUs), Tensor Processing Units (TPUs), and neuromorphic chips, which consume significantly less power compared to traditional hardware [\cite{clemm2024greenaicurrentstatus}, \cite{tabbakh2024}]. Advances in hardware design are now targeting the creation of AI accelerators that combine high performance with environmental sustainability. This effort includes developing processors that can run AI algorithms with very low energy consumption [\cite{Rahmani2023}]. Ongoing research aims to create AI devices capable of harvesting energy from their surroundings to power AI systems [\cite{BOLONCANEDO2024128096}]. Examples include utilizing ambient light, vibrations, and heat. By incorporating energy-harvesting features, these AI devices could become more self-sufficient, reducing reliance on external power sources and promoting a more sustainable deployment of AI [\cite{DIVYA2023108084}, \cite{ALI2023101124}].

Neuromorphic computing represents a significant shift in computing technology, inspired by the architecture of the human brain to create more efficient and adaptive systems. The human brain's remarkable energy efficiency allows it to perform complex tasks with far less energy than conventional computers. By emulating this efficiency, neuromorphic computing aims to reduce the energy consumption of computational processes to levels similar to biological brains. This is achieved through the use of spiking neural networks (SNNs), which mimic the behavior of biological neurons, leading to substantial energy savings and performance enhancements, especially in AI applications that require real-time processing and decision-making. Additionally, a sustainability proposal by \cite{Oh2023} suggests creating biodegradable neural network hardware to mitigate the impact of technological waste.

Powering data centers with renewable energy sources is a crucial step towards reducing the carbon footprint of the rapidly growing digital infrastructure. Integrating renewable energy, such as solar and wind power, into data center operations can help mitigate the environmental impact of these energy-intensive facilities [\cite{brownandgorham2014}]. However, the variability of renewable energy sources poses challenges, which can be addressed through innovative solutions like virtual batteries that shift computational demand to match power availability [\cite{agarwal2021}]. Additionally, advancements in energy storage technologies and smart grid integration are essential for ensuring a stable and reliable power supply [\cite{hussain2024}]. These efforts not only support sustainability goals but also align with the increasing corporate commitments to carbon neutrality [\cite{agarwal2021}].

CCS is a versatile technology and can be an enabler for data center decarbonization, with the potential to capture 90-95+ percent of a power generation facility’s CO\textsubscript{2} emissions [\cite{Barlow2025}]. A business model is emerging which applies CCS to purpose-built power generation facilities "behind the meter" (not connected to municipal electric grids) to minimize data center carbon emissions while meeting their firm, high-reliability power demand. In this model, CO\textsubscript{2} emissions from natural gas turbines are captured, transported, and permanently stored before they reach the atmosphere – significantly decarbonizing the data center power supply. Industry is responding to the data center decarbonization opportunity in the U.S. and other countries. Further details can be found in [\cite{GlobalCCSInstitute2025}].

Recent advancements in deep learning have focused on reducing the size of neural networks to enhance computational efficiency and enable deployment in resource-constrained environments. One approach involves a pruning algorithm that retrains the smaller, pruned model at its faster, initial learning rate, resulting in models that are simpler and more accurate [\cite{renda_2020}]. Another comprehensive review highlights techniques such as pruning, quantization, low-rank factorization, knowledge distillation, and transfer learning, all of which effectively reduce model size while retaining performance. A study on global weight compression demonstrates that deploying deep neural networks to devices requiring real-time processing can save memory, reduce storage size, and lower computational requirements [\cite{dantas_2024}]. An example advancement is the PocketNet paradigm, which aims to reduce the size of deep learning models used in medical image analysis. The authors propose a modification to existing convolutional neural network architectures that limits the growth of the number of channels, thereby significantly reducing the number of parameters. This approach maintains comparable performance to conventional neural networks while using up to 90 percent less GPU memory and speeding up training times by up to 40 percent. The paper demonstrates the effectiveness of PocketNet in various segmentation and classification tasks, highlighting its potential for deployment in resource-constrained settings [\cite{celaya_2024}]. 

DeepSeek, a Chinese AI model, has garnered significant attention for its energy-efficient approach to artificial intelligence. Unlike traditional models, DeepSeek employs a "mixture of experts" technique during training, where only a portion of the model's parameters are activated at any given time. This method, combined with improved reinforcement learning techniques, has led to more efficient training processes. However, while DeepSeek's training phase is notably energy-efficient, its inference phase tends to be more energy-intensive. The model generates longer responses compared to similar models, which offsets some of the energy savings achieved during training [\cite{odonnell2025}]. Despite its energy-efficient training, DeepSeek's overall impact on energy consumption remains complex. The model's longer inference responses require substantial computational power, which can lead to higher energy usage in data centers. This phenomenon is consistent with the broader trend in AI development, where efficiency gains in training often result in increased overall energy consumption due to the expanded capabilities and applications of the models [\cite{odonnell2025}]. As companies strive to develop more intelligent systems, they tend to invest more in training, thereby using more energy. This dynamic underscores the importance of considering both training and inference phases when evaluating the energy efficiency of AI models. The introduction of DeepSeek has significant implications for the tech industry and energy sector. While its energy-efficient training methods offer a promising solution to the growing energy demands of AI, the model's inference phase highlights the need for balanced approaches to AI development. The rapid adoption of DeepSeek-like models could strain power grids and challenge energy security.

To ensure AI remains an environmentally friendly tool, several strategies are crucial: These include focusing on green AI hardware, and powering data centers with renewable energy sources to reduce the environmental impact of AI operations. Research into model optimization techniques can also lead to more energy-efficient models as discussed above. Furthermore, standardized environmental impact assessments and collaboration between AI developers and sustainability experts can promote transparency and responsible development throughout the AI lifecycle.

\section{Conclusions}

In this paper, we analyzed the impact of the rise of AI, the demand for building data centers, electricity consumption and demand patterns, and the subsequent effects on CO\textsubscript{2} emissions and net zero goals. Specifically, we presented a comprehensive scenario of various factors driving electricity consumption, outlining both near-term and long-term scenarios, and identifying factors to mitigate these impacts. We addressed the critical question of whether AI will result in net positive or net negative emissions from both short-term and long-term perspectives.

We provided a probabilistic assessment of CO\textsubscript{2} emissions based on the mean scenario derived from recent projections by the \cite{iea2025, paccouandwijnhoven2024} using the methodology presented in \cite{Shell_Sce_2025} . Our P5, P50, and P95 cases indicate that while AI growth is expected to continue over the next ten years, it is unlikely to reach extreme scenarios depicted in \cite{iea2025, paccouandwijnhoven2024}. Instead, we anticipate deviations from the mean scenario, but not to such extreme extents. This is because most emissions are expected to come from data centers, and the growth in data centers does not support such extreme scenarios. Additionally, we do not foresee an energy crisis scenario, as there has been no significant precedent for such an event in the past 20 years, including during the COVID-19 pandemic. 

In conclusion, AI presents both a challenge and an opportunity for achieving net-zero emissions. While the near-term increases in energy demand and emissions are concerning, the long-term potential for emissions reductions in other sectors is promising. It is critical to focus on responsible AI development and deployment to maximize its benefits for the net-zero transition. This includes leveraging public and private funds for further research, collaboration, and investments in sustainable AI developments and practices. 

\section{Acknowledgements}
The authors thank Shell Information Technology International Inc. and Shell International Exploration and Production Inc. for permission to publish this work. Additionally, we are grateful to our colleagues—including Juben Chheda, Mariela Araujo, Christian Davies, Pak Leung, and Albena Mateeva—for their meaningful discussions and valuable contributions that shaped the final paper.
\bibliographystyle{unsrtnat}
\bibliography{references}  

\appendix


\begin{table}[t]
\caption{The anticipated electricity consumption for various tasks related to AI chatbot models [\cite{Luccioni20240605}]. As shown, text-based tasks are significantly less demanding in terms of electricity consumption, while image-based tasks require substantially more energy. For context, generating 1,000 words of text is comparable to charging a smartphone battery by 9\%. In contrast, completing image-based tasks can consume as much energy as charging approximately 522 smartphones.}
\label{tab:inference}
\begin{tabular}{p{7.3cm}p{4.1cm}}
\toprule 
\textbf{Task} & \textbf{Inference energy consumption (Wh)} \vspace{0.05cm} \\  \toprule 
Text classification   & 2 \\ \hline
      Text generation & 47 \\ \hline
      Summarization & 49 \\ \hline
      Object detection & 38 \\ \hline
      Image captioning & 63\\ \hline
      Image generation & 2907 \\ \hline
\end{tabular}
\end{table}

\begin{table}[t]
\caption{The list of key variables potentially impacting the CO\textsubscript{2} emissions at data centers include the following factors: electricity consumption by the semiconductor industry [\cite{greenpeacereport2024}], electricity consumption by data centers [\cite{paccouandwijnhoven2024}], the energy mix factor indicating the percentage of fossil feedstock-based electricity generation, the percentage of AI workloads at the data centers, and CO\textsubscript{2} emissions in Mt per year. The data from Figure~\ref{fif:co2foecast_scenarios}b were used to find the mean scenario and the corresponding numbers.} \label{tab:key_variables}
\begin{tabular}{lcccccc}
\toprule 
\thead{Year} & \thead{Electricity consumption by \\ semiconductor industry (TWh)} & 
\thead{Electricity consumption by \\ data centers (TWh)} & \thead{Energy mix factor} & 
\thead{AI workload \\ at data centers} & \thead{CO\textsubscript{2} emissions \\ (Mt/year)} \\  \toprule 
    2020 &  91 & 269 & 0.62 & 0.02 & 1.03 \\ \hline
    2021 & 101 & 300 & 0.59 & 0.03 & 2.07  \\ \hline
    2022 & 112 & 330 & 0.59 & 0.06 & 4.12 \\ \hline
    2023 & 126 & 361 & 0.56 & 0.11 & 8.18  \\ \hline
    2024 & 140 & 416 & 0.55 & 0.17 & 14.18 \\ \hline
    2025 & 156 & 482.5 & 0.53 & 0.2 & 20.08\\ \hline
    2026 & 173 & 550 & 0.52 &  0.3 & 32.7                         \\ \hline
    2027 & 191 & 650 & 0.50 &  0.43 & 54.7                        \\ \hline
    2028 & 210 & 746 & 0.47 &  0.55 & 80                        \\ \hline
    2029 & 229 & 838 & 0.45 &  0.65 & 104.5                        \\ \hline
    2030 & 240 & 918 & 0.43 &  0.73 & 126                         \\ \hline
    2031 & 245 & 981 & 0.41 &  0.69 & 124                         \\ \hline
    2032 & 266 & 1038 & 0.39 & 0.67 & 124                          \\ \hline
    2033 & 288 & 1080 & 0.37 & 0.65 & 124                         \\ \hline
    2034 & 310 & 1118 & 0.35 & 0.63 & 121                          \\ \hline
    2035 & 334 & 1148 & 0.33 & 0.64 & 123                         \\ \hline
\end{tabular}
\end{table}

\end{document}